%% file: main.tex
\documentclass[sigconf]{acmart}
\usepackage{pifont}
\usepackage{multirow}
\usepackage{soul}
\usepackage{algorithm}
\usepackage{algorithmic}
\usepackage{enumitem}
\AtBeginDocument{%
  }

\copyrightyear{2026}
\acmYear{2026}
\setcopyright{cc}
\setcctype{by}
\acmConference[CIKM '26]{Proceedings of the 35th ACM International Conference on Information and Knowledge Management}{November 7--11, 2026}{Rome, Italy.}
\acmBooktitle{Proceedings of the 35th ACM International Conference on Information and Knowledge Management (CIKM '26), November 7--11, 2026, Rome, Italy}
\acmISBN{979-8-4007-2539-5/2026/11}
\acmDOI{10.1145/3799682.3840707}
\begin{document}

\title{RiskTraf: Risk-Extrapolated Residual Learning for Multi-Variate Traffic Flow Prediction}

\author{Guangyu Wang}
\email{hulnegy@gmail.com}
\affiliation{%
  \institution{Dongbei University of Finance \& Economics}
  \city{Dalian}
  \state{Liaoning}
  \country{China}
}

\author{Zhidan Liu}
\authornote{Corresponding author.}
\email{zhidanliu@hkust-gz.edu.cn}
\affiliation{%
  \institution{Hong Kong University of Science and Technology (Guangzhou)}
  \city{Guangzhou}
  \state{Guangdong}
  \country{China}
}
\renewcommand{\shortauthors}{Guangyu Wang and Zhidan Liu}

\begin{abstract}
Traffic sensors commonly record flow, speed, and occupancy, but standard traffic flow forecasting benchmarks and models rarely exploit all three raw measurements reliably. Although speed and occupancy provide sensor-native traffic-state information beyond flow alone, existing releases often omit these variables, replace them with proxies, or contain logically inconsistent records. Moreover, direct empirical risk minimization over three-variable inputs may exploit regime-dependent shortcuts, as the relationships among flow, speed, and occupancy vary substantially between free-flow and congested states. We introduce \textbf{PEMSB-3V}, a public benchmark suite that preserves raw flow, speed, and occupancy measurements from PeMS detectors for flow prediction. We also propose \textbf{RiskTraf}, a model-agnostic risk-extrapolated residual plug-in. For each trained spatio-temporal backbone, RiskTraf freezes the selected checkpoint and learns a lightweight zero-start residual head from historical speed and occupancy. The residual head constructs ordered traffic-risk environments and optimizes horizon-wise flow corrections with a risk extrapolation objective, thereby mitigating regime-specific shortcut correlations without modifying the backbone. Extensive experiments demonstrate that RiskTraf consistently improves diverse forecasting backbones and outperforms debiasing and distribution-shift adaptation methods. Our code and benchmark are available at \url{https://github.com/Guangyu4/RiskTraf}.
\end{abstract}




\maketitle

\input{content/1_introduction}

\input{content/2_benchmark}

\input{content/3_method}

\input{content/4_experiments}

\input{content/2_related_work}
\input{content/7_conclusion}

\begin{acks}
This work was supported in part by National Natural Science Foundations of China under Grant No.~62572416 and the Guangdong Provincial Key Lab of Integrated Communication, Sensing and Computation for Ubiquitous Internet of Things under Grant No.~2023B1212010007.
\end{acks}

\section*{GenAI Usage Disclosure}
Generative AI tools were used only to assist with language polishing, wording refinement, and formatting during manuscript preparation. The research ideas, method design, experiments, analyses, figures, tables, and conclusions were produced and verified by the authors, who take full responsibility for the content of this paper.



\bibliographystyle{ACM-Reference-Format}
\bibliography{sample-base}
 


\end{document}

%% file: content/1_introduction.tex
\section{Introduction} \label{sec:intro}


Traffic flow prediction aims to forecast the number of vehicles passing through road segment stations over future time intervals based on historical observations. In widely used benchmarks such as PeMS (Freeway Performance Measurement System)~\cite{chen2001pems}, traffic flow is typically defined as the vehicle count recorded by loop detectors within fixed time windows, e.g., every 5 minutes. As a fundamental indicator of road utilization, accurate flow prediction plays a critical role in intelligent transportation systems~\cite{wang2026stpia}, enabling traffic authorities to anticipate congestion, optimize signal control, and allocate transportation resources more effectively~\cite{liu2022dispatch}.



Recent studies have made substantial progress in traffic forecasting by designing increasingly expressive spatio-temporal models, including graph neural networks~\cite{li2018dcrnn,wu2020mtgnn} and attention-based architectures~\cite{liu2023stae}. However, as the field matures, performance gains from purely architectural innovations have become increasingly marginal~\cite{shao2024benchmark}, while model complexity and computational costs continue to grow. This trend has motivated researchers to exploit richer contextual information for more accurate prediction. Existing efforts have incorporated external sources such as weather conditions~\cite{dunne2013weather,lee2015weather} and textual information~\cite{zhang2024chattraffic}. Nevertheless, incorporating such auxiliary sources often requires additional collection, storage, and maintenance, thereby increasing the data-side overhead of prediction systems~\cite{chen2025expand}. More importantly, they are difficult to align with traffic measurements: weather typically affects broad regions while sensors observe localized road segments~\cite{zhang2017stresnet}, and textual signals such as news are updated at much coarser temporal granularity than traffic sensors~\cite{he2013twitter}. Such spatial and temporal mismatches may introduce additional noise during data fusion.

Compared with external auxiliary sources, a more natural way to enrich traffic representations is to leverage multiple variables reported by the same traffic sensor. In addition to flow, loop detectors commonly record \textit{speed} and \textit{occupancy}, which describe complementary aspects of traffic states. Speed captures vehicle movement dynamics, while occupancy measures the fraction of time during which detectors are occupied by vehicles. Since these variables are collected from the same sensors and at the same temporal resolution as flow, they provide sensor-native auxiliary information without the spatial and temporal misalignment of external data. From an information-theoretic perspective~\cite{ash2012information}, jointly modeling heterogeneous sensor variables can provide additional state information beyond any single variable. A few studies have therefore begun to exploit flow, speed, and occupancy for traffic forecasting~\cite{ji2022stden,shao2025physics,lu2023physics}.

However, incorporating speed and occupancy into deep forecasting models is not a straightforward extension of flow-only prediction. Their effective use requires satisfying two coupled requirements: obtaining reliable three-variable measurements from the same set of sensors, and learning from these auxiliary variables without relying on regime-specific shortcuts.

\textbf{\textit{Challenge 1: Lack of reliable multi-variate benchmarks.}} 
Existing traffic sensor releases do not guarantee that flow, speed, and occupancy are simultaneously valid and usable. Raw loop-detector records often require strict screening based on validity thresholds and traffic-flow consistency~\cite{turochy2000new}, and field detectors may suffer from faults such as stuck-off, stuck-on, or hanging-on behaviors~\cite{chen2003detecting}. Moreover, speed and occupancy are sensitive to detector configurations, lane definitions, loop lengths, and road types. As a result, a sensor that provides usable flow measurements may still be unreliable for raw three-variable modeling. This makes it difficult to systematically evaluate methods that exploit flow, speed, and occupancy under a standardized benchmark setting.

\textbf{\textit{Challenge 2: Regime-dependent auxiliary correlations.}} 
Although speed and occupancy contain useful traffic-state information, their relationships with flow are not invariant across traffic regimes. For example, under congested conditions, slower vehicles may occupy loop detectors for longer durations, increasing occupancy even when the observed flow is similar. Consequently, naively treating speed and occupancy as ordinary covariates may cause models to exploit regime-specific shortcuts rather than robust predictive patterns. Models trained with empirical risk minimization (ERM) can overfit such spurious correlations, leading to degraded generalization across traffic states. Simply discarding speed and occupancy avoids this risk but also wastes valuable sensor information. Existing distribution-shift adaptation~\cite{chen2021trafficstream,fan2023dishts} and debiasing methods~\cite{zhou2023caustg} partially address related issues, but they often rely on specialized shift or bias detectors, whose effectiveness is limited by the accuracy and availability of such detectors.

To address these challenges, we introduce both a reliable benchmark suite and a risk-aware learning framework. First, we construct the \textbf{PEMSB-3V Benchmark} consisting of four datasets: PEMS03-B, PEMS04-B, PEMS07-B, and PEMS08-B. The benchmark is district-aligned with widely used PeMS datasets while preserving raw flow, speed, and occupancy measurements reported by PeMS detectors. By conducting sensor-level validation, PEMSB-3V provides a standardized testbed for studying historical three-variable inputs with flow-only forecasting targets.

Second, we propose \textbf{RiskTraf}, a \underline{\textbf{risk}} extrapolation plug-in for \underline{\textbf{traf}}fic prediction. Rather than using speed and occupancy as unrestricted covariates or replacing existing forecasting architectures, RiskTraf treats them as historical auxiliary signals for residual correction. Specifically, given a trained three-variable backbone, we freeze its parameters and attach a lightweight zero-initialized residual head driven by historical speed and occupancy. The residual head learns horizon-wise flow corrections across ordered low-to-high traffic-risk environments, encouraging robust improvements while reducing reliance on regime-specific auxiliary correlations. RiskTraf performs rollback when validation MAE does not improve.

The main contributions of this work are summarized as follows:
\begin{itemize}
    \item We release \textbf{PEMSB-3V}, a district-aligned benchmark suite that enables standardized evaluation of flow prediction with historical flow, speed, and occupancy measurements.
    
    \item We formulate regime-dependent auxiliary correlations as a key obstacle in multi-variate traffic forecasting and propose \textbf{RiskTraf}, a REx-based residual plug-in that improves trained backbones without replacing their architectures.
    
    \item We conduct extensive experiments on multiple PEMSB-3V datasets and diverse spatio-temporal backbones, showing that RiskTraf achieves consistent improvements over strong baselines, debiasing methods, and distribution-shift adaptation approaches.
\end{itemize}

The rest of this paper is organized as follows. Section~\ref{sec:benchmark} introduces the PEMSB-3V benchmark. Section~\ref{sec:methodology} presents the RiskTraf framework. Section~\ref{sec:experiments} reports experimental results and analyses. Section~\ref{sec:related} discusses related work, and Section~\ref{sec:conclusion} concludes the paper.

%% file: content/2_benchmark.tex
\section{PEMSB-3V Benchmark}
\label{sec:benchmark}

We construct \textbf{PEMSB-3V}, a benchmark suite consisting of four district-level datasets: \textit{PEMS03-B}, \textit{PEMS04-B}, \textit{PEMS07-B}, and \textit{PEMS08-B}. The suite is built from the California Department of Transportation Performance Measurement System (PeMS)\footnote{\url{https://pems.dot.ca.gov/}} and its official source page\footnote{\url{https://dot.ca.gov/programs/traffic-operations/mpr/pems-source}}. PeMS is described by Caltrans as a statewide freeway monitoring system with nearly 40,000 detectors. 
Unlike existing releases that replace missing variables with proxies or temporal codes, PEMSB-3V retains only detectors whose native PeMS records contain all three raw measurements: \textit{flow}, \textit{speed}, and \textit{occupancy}. The overall construction pipeline is summarized in Algorithm~\ref{alg:benchmark}.


\begin{algorithm}[H]
\caption{PEMSB-3V Benchmark Construction Pipeline}
\label{alg:benchmark}
\begin{algorithmic}[1]
\small
\REQUIRE District ID, time range $[t_{\text{start}}, t_{\text{end}}]$, completeness threshold $\rho$
\ENSURE Multi-variate traffic benchmark $\mathcal{D}$ and adjacency matrix $\mathbf{A}$
\STATE Query detector metadata from PeMS
\STATE Keep only detectors with raw flow, speed, and occupancy channels
\STATE Filter detectors whose temporal completeness is below $\rho$
\FOR{each day in $[t_{\text{start}}, t_{\text{end}}]$}
    \STATE Download 5-minute flow, speed, and occupancy records
    \STATE Remove malformed timestamps and interpolate only short gaps
\ENDFOR
\STATE Aggregate the cleaned records into $\mathbf{X} \in \mathbb{R}^{T \times N \times 3}$
\STATE Construct the road topology and compute pairwise distances
\STATE Build $\mathbf{A}$ with a distance kernel and sparsification threshold
\STATE Split the benchmark into train/val/test by ratio 6:2:2
\RETURN $\mathcal{D}$, $\mathbf{A}$
\end{algorithmic}
\end{algorithm}

\begin{figure}[t]
    \centering
    \includegraphics[width=0.8\linewidth]{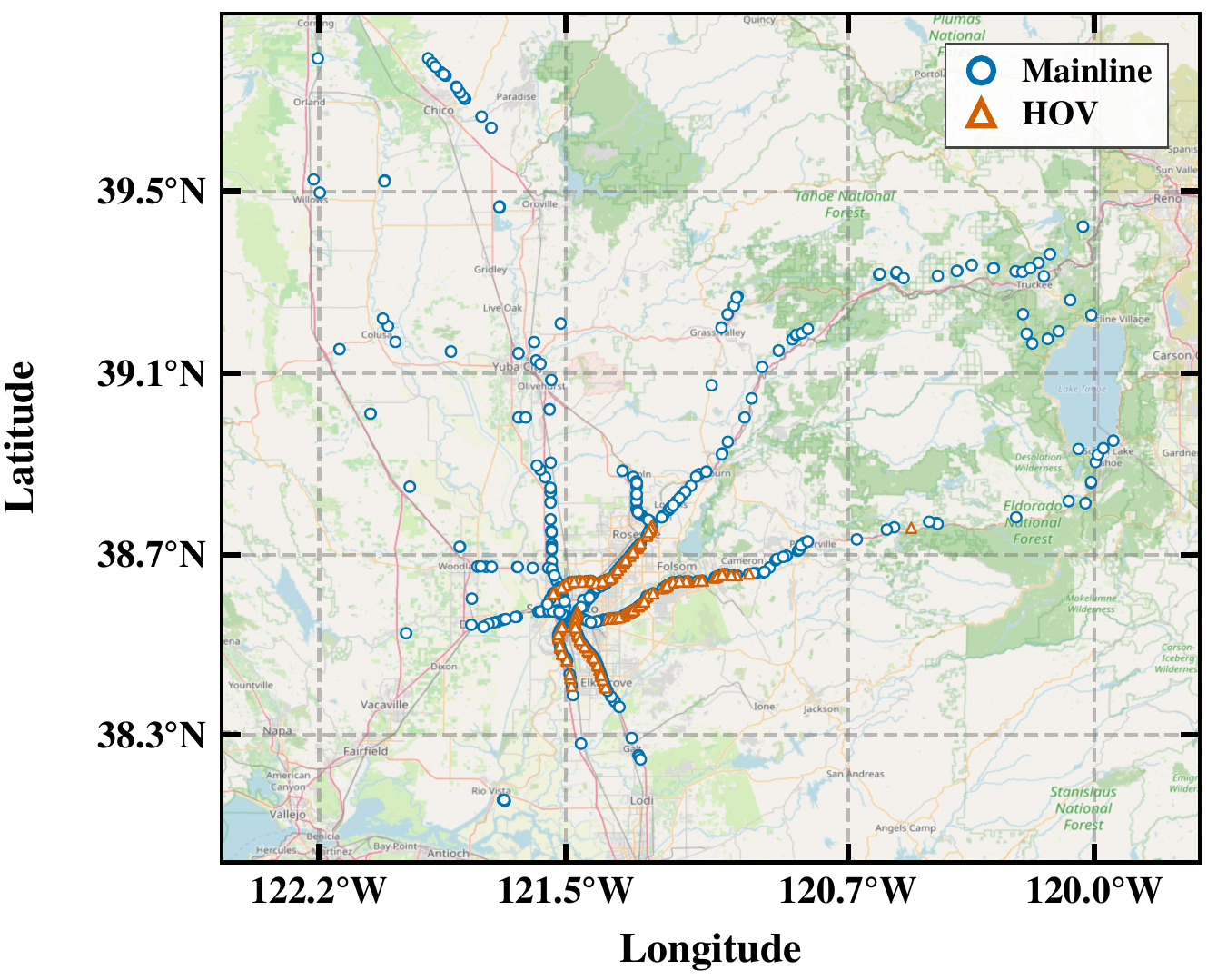}
    \caption{High-quality PEMS03-B sensors after metadata-based screening. The map shows 1,013 merged mainline/HOV detectors with complete metadata.}
    \label{fig:metadata_quality_screening}
    \Description{Map of high-quality PEMS03-B freeway mainline and HOV detector locations after metadata-based screening.}
\end{figure}

To ensure that the benchmark is built from sensors with physically interpretable three-variable measurements, we apply a metadata-based progressive screening procedure before constructing the final sensor set. The screening follows the earliest-failed-check rule. 
We first perform an ID coverage check. A target sensor without a corresponding PeMS metadata record is marked as \textit{Missing Metadata}, since its static information cannot be verified. We then perform a spatial localization check. If latitude or longitude is missing, invalid, or outside the legal geographic range, the detector is marked as \textit{Missing Geolocation} and excluded from map-based topology construction. Next, we check static structural completeness using detector length and lane count. Records with empty, non-numeric, or non-positive values are marked as \textit{Invalid Static Attributes}, because flow, speed, and occupancy depend on the detector segment and lane definition. Finally, we check road-type consistency. Detectors with $\mathrm{Type}\in\{\mathrm{ML}, \mathrm{HV}\}$ are retained as freeway mainline or HOV sensors, whereas $\mathrm{Type}\in\{\mathrm{OR}, \mathrm{FR}, \mathrm{FF}\}$ are treated as ramp or connector sensors and excluded from the mainline subset.


Figure~\ref{fig:metadata_quality_screening} illustrates the resulting PEMS03-B sensor footprint after metadata-based screening. Among 1,708 target sensor IDs, 41 are removed due to missing metadata and 2 due to invalid or missing geolocation. Among the remaining 1,665 geolocated records, 1,039 pass all metadata checks as complete mainline/HOV sensors, while 626 are identified as ramp or connector sensors. After merging duplicate records with the same location, road direction, and detector type, the final map contains 1,013 high-quality points, including 762 mainline and 251 HOV points. This procedure makes each discarded sensor traceable to an interpretable failure category, rather than removing sensors through opaque filtering.

The subset design follows the administrative organization of PeMS rather than an arbitrary partition. Specifically, PEMS03-B, PEMS04-B, PEMS07-B, and PEMS08-B correspond to Caltrans districts 03, 04, 07, and 08, respectively\footnote{\url{https://cwwp2.dot.ca.gov/documentation/district-map-county-chart.htm}}. This preserves geographic provenance while maintaining comparability with widely used PeMS benchmarks.


\input{tables/benchmarks}

We further audit representative public traffic benchmarks to examine whether they support the raw flow-speed-occupancy forecasting setting considered in this paper. As shown in Table~\ref{table:benchmarks}, most existing benchmarks either omit speed and occupancy, replace them with proxy variables, or contain logically inconsistent records. For example, PEMS03 and PEMS07 use daily and weekly temporal encodings in place of raw speed and occupancy, while PEMS04 and PEMS08 include physically implausible speed records, such as positive speed when flow is zero and truncated low-speed ranges. These issues prevent existing benchmarks from serving as reliable testbeds for studying how raw speed and occupancy contribute to flow prediction.

%% file: tables/benchmarks.tex
\begin{table}[tb]
    \caption{Availability and usability of raw flow, speed, and occupancy measurements in representative traffic benchmarks. \textcolor{green}{\checkmark}: usable raw measurement; \textcolor{red}{\ding{55}}: not provided; \textcolor{red}{$\bigcirc$}: provided but logically inconsistent.}
    \label{table:benchmarks}
   \vspace{-1em}
    
    \centering
    \resizebox{0.48\textwidth}{!}{
    \begin{tabular}{@{} c c c c c c c c @{}}
    \toprule
    \textbf{Source} & \textbf{Benchmark} & \textbf{Nodes} & \textbf{Interval} & \textbf{Flow} & \textbf{Speed} & \textbf{Occupancy} \\ \midrule
    \multirow{2}{*}{DCRNN~\citep{li2018dcrnn}}  & METR-LA    & 207   & 5 mins   & \textcolor{red}{\ding{55}} & \textcolor{green}{\checkmark} & \textcolor{red}{\ding{55}} \\
                            & PEMS-BAY   & 325   & 5 mins   & \textcolor{red}{\ding{55}} & \textcolor{green}{\checkmark} & \textcolor{red}{\ding{55}} \\ 
    LSTNet\citep{lai2018modeling}                  & Traffic    & 862   & 1 hour   & \textcolor{green}{\checkmark} & \textcolor{red}{$\bigcirc$} & \textcolor{red}{$\bigcirc$} \\
    \multirow{2}{*}{STGCN\citep{yu2018spatio}}  & PEMSD7(M)  & 228   & 5 mins   & \textcolor{green}{\checkmark} & \textcolor{red}{$\bigcirc$} & \textcolor{red}{$\bigcirc$} \\
                            & PEMSD7(L)  & 1,026 & 5 mins   & \textcolor{green}{\checkmark} &  \textcolor{red}{$\bigcirc$} & \textcolor{red}{$\bigcirc$} \\
    \multirow{2}{*}{ASTGCN\citep{guo2019attention}} & PEMSD4-I   & 228   & 5 mins   & \textcolor{green}{\checkmark} & \textcolor{red}{$\bigcirc$} & \textcolor{red}{$\bigcirc$} \\
                            & PEMSD8-I   & 1,979 & 5 mins   & \textcolor{green}{\checkmark} & \textcolor{red}{$\bigcirc$} & \textcolor{red}{$\bigcirc$} \\
    \multirow{4}{*}{STSGCN\citep{Song_Lin_Guo_Wan_2020}} & PEMS03     & 358   & 5 mins   & \textcolor{green}{\checkmark} &  \textcolor{red}{$\bigcirc$} & \textcolor{red}{$\bigcirc$}\\
                            & PEMS04     & 307   & 5 mins   & \textcolor{green}{\checkmark} &  \textcolor{red}{$\bigcirc$} & \textcolor{red}{$\bigcirc$} \\
                            & PEMS07     & 883   & 5 mins   & \textcolor{green}{\checkmark} &  \textcolor{red}{$\bigcirc$} & \textcolor{red}{$\bigcirc$}\\
                            & PEMS08     & 170   & 5 mins   & \textcolor{green}{\checkmark} &  \textcolor{red}{$\bigcirc$} & \textcolor{red}{$\bigcirc$} \\
    \multirow{4}{*}{LargeST\citep{liu2023largest}} & CA       & 8,600 & 5 mins   & \textcolor{green}{\checkmark} & \textcolor{red}{\ding{55}} & \textcolor{red}{\ding{55}} \\
                            & GLA        & 3,834 & 5 mins   & \textcolor{green}{\checkmark} & \textcolor{red}{\ding{55}} & \textcolor{red}{\ding{55}} \\
                            & GBA        & 2,352 & 5 mins   & \textcolor{green}{\checkmark} & \textcolor{red}{\ding{55}} & \textcolor{red}{\ding{55}} \\
                            & SD         & 716   & 5 mins   & \textcolor{green}{\checkmark} & \textcolor{red}{\ding{55}} & \textcolor{red}{\ding{55}} \\ 
    \multirow{11}{*}{XXLTraffic\citep{yinXXLTrafficExpandingExtremely2025}} 
                            & Full\_PEMS03 & 1,809 & 5 mins & \textcolor{green}{\checkmark} & \textcolor{red}{$\bigcirc$} & \textcolor{red}{$\bigcirc$} \\ 
                            & Full\_PEMS04 & 4,089 & 5 mins & \textcolor{green}{\checkmark} & \textcolor{red}{\ding{55}} & \textcolor{red}{\ding{55}} \\ 
                            & Full\_PEMS05 & 573   & 5 mins & \textcolor{green}{\checkmark} & \textcolor{red}{\ding{55}} & \textcolor{red}{\ding{55}} \\ 
                            & Full\_PEMS06 & 705   & 5 mins & \textcolor{green}{\checkmark} & \textcolor{red}{\ding{55}} & \textcolor{red}{\ding{55}} \\ 
                            & Full\_PEMS07 & 4,888 & 5 mins & \textcolor{green}{\checkmark}& \textcolor{red}{\ding{55}} & \textcolor{red}{\ding{55}} \\ 
                            & Full\_PEMS08 & 2,059 & 5 mins & \textcolor{green}{\checkmark} & \textcolor{red}{\ding{55}} & \textcolor{red}{\ding{55}} \\ 
                            & Full\_PEMS10 & 1,378 & 5 mins & \textcolor{green}{\checkmark}& \textcolor{red}{\ding{55}} & \textcolor{red}{\ding{55}} \\ 
                            & Full\_PEMS11 & 1,440 & 5 mins & \textcolor{green}{\checkmark} & \textcolor{red}{\ding{55}} & \textcolor{red}{\ding{55}} \\ 
                            & Full\_PEMS12 & 2,587 & 5 mins & \textcolor{green}{\checkmark} & \textcolor{red}{\ding{55}} & \textcolor{red}{\ding{55}} \\ 
                            & tfNSW       & 27    & 60 mins & \textcolor{green}{\checkmark} & \textcolor{red}{\ding{55}} & \textcolor{red}{\ding{55}} \\ \midrule
    TraffiDent~\citep{gou2026traffident} & TraffiDent$^{*}$ & 16,972 & 5 mins & \textcolor{green}{\checkmark} & \textcolor{green}{\checkmark} & \textcolor{green}{\checkmark} \\ \midrule
    \multirow{4}{*}{PEMSB-3V} & PEMS03-B     & 1,013 & 5 mins   & \textcolor{green}{\checkmark} & \textcolor{green}{\checkmark} & \textcolor{green}{\checkmark} \\
                            & PEMS04-B    & 2,474 & 5 mins   & \textcolor{green}{\checkmark} & \textcolor{green}{\checkmark} & \textcolor{green}{\checkmark} \\
                            & PEMS07-B    & 2,788 & 5 mins   & \textcolor{green}{\checkmark} & \textcolor{green}{\checkmark}& \textcolor{green}{\checkmark} \\
                            & PEMS08-B    & 1,515   & 5 mins   & \textcolor{green}{\checkmark} & \textcolor{green}{\checkmark}& \textcolor{green}{\checkmark} \\  
    \bottomrule
    \end{tabular}}
    \vspace{0.5em}
    \footnotesize{$^{*}$TraffiDent provides all three variables but does not apply sensor-type filtering or provide sufficient metadata to distinguish detector types.}
    
    \end{table}

%% file: content/3_method.tex
\section{Methodology}
\label{sec:methodology}

\begin{figure*}[t]
    \centering
    \includegraphics[width=0.95\textwidth]{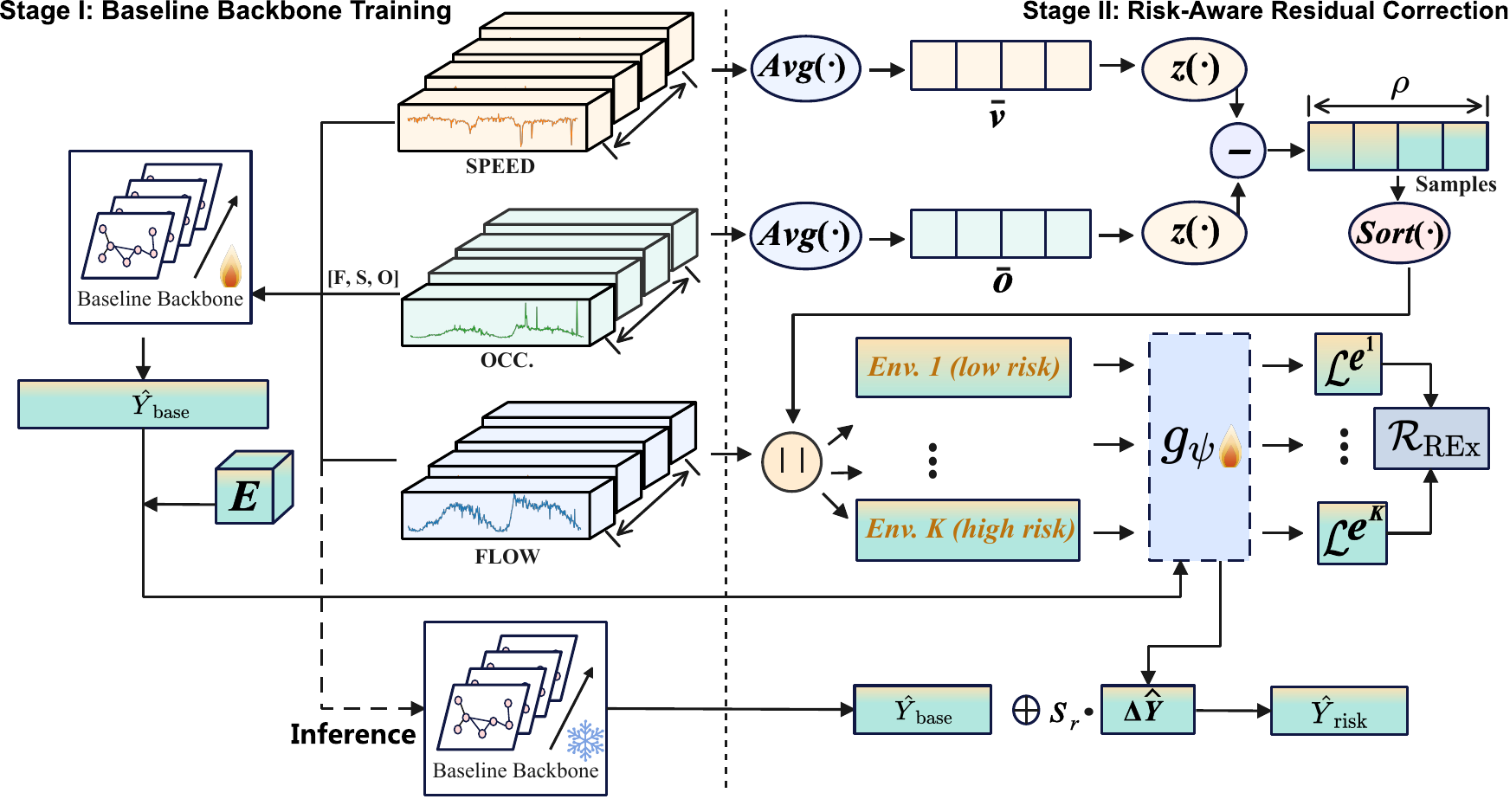}
    \caption{Overview of RiskTraf. Stage I trains a backbone to predict future flow from historical flow, speed, and occupancy. Stage II freezes the backbone, constructs traffic-risk environments from historical speed and occupancy, and learns a lightweight residual head with risk extrapolation to correct the baseline prediction.}
    \label{fig:method}
\end{figure*}

In this section, we present \textbf{RiskTraf}, a model-agnostic risk-extrapolation plug-in for multi-variate traffic forecasting. As illustrated in Figure~\ref{fig:method}, RiskTraf adopts a paired two-stage design without modifying the forecasting backbone:
\begin{itemize} [leftmargin=1.2em, itemsep=2pt, topsep=2pt]
    \item \textbf{Stage I} trains a given standard spatio-temporal backbone with 12 historical steps of traffic flow, speed, and occupancy to predict future flow.
    
    \item \textbf{Stage II} freezes the trained backbone and attaches a lightweight residual head driven by historical speed and occupancy. The residual head is optimized with a risk extrapolation objective to learn robust flow corrections across traffic-risk environments.
\end{itemize}

\subsection{Problem Formulation and Paired Protocol}

Let $F$, $S$, and $O$ denote traffic flow, speed, and occupancy, respectively. For a sample ending at time $t-1$, the historical input is
\begin{equation}
X_t = \{F_{t-T:t-1}, S_{t-T:t-1}, O_{t-T:t-1}\},
\end{equation}
where $T$ is the input length and each variable is observed over all sensor nodes. The prediction target is future flow only:
\begin{equation}
Y_t = F_{t:t+H-1},
\end{equation}
where $H$ is the prediction horizon. Thus, speed and occupancy are available only as historical auxiliary measurements.

RiskTraf is applied under a paired protocol for each dataset-backbone pair. Let $B_\phi$ be an arbitrary spatio-temporal backbone. The Stage-I baseline prediction is given by
\begin{equation}
\hat{Y}_{\text{base}} = B_\phi(X_t).
\end{equation}
After selecting the baseline checkpoint by validation MAE, RiskTraf starts from that checkpoint, freezes the backbone parameters $\phi$, and learns a residual correction head $g_\psi$ from historical speed and occupancy:
\begin{equation}
\hat{Y}_{\text{risk}} = \hat{Y}_{\text{base}} + s_r \cdot g_\psi(S_{t-T:t-1}, O_{t-T:t-1}),
\end{equation}
where $s_r$ is a small residual scaling factor; during Stage II only the residual parameters $\psi$ are updated.

\subsection{Stage I: Baseline Backbone Training}

The first stage follows the standard empirical risk minimization (ERM) setting adopted by existing traffic forecasting models. 
Given a historical three-variable sequence $X$, the backbone $B_\phi$ predicts the future flow sequence $Y_t$. The optimization objective is
\begin{equation}
\min_{\phi} \; \mathbb{E}_{(X,Y)} \left[\ell(B_\phi(X), Y)\right],
\end{equation}
where $\ell$ is the masked mean absolute error (MAE) computed on inverse-scaled flow values. The checkpoint with the best validation MAE is selected as the reference checkpoint for both the vanilla backbone and the subsequent RiskTraf stage.



\subsection{Stage II: Risk-Aware Residual Plug-in}
\label{sub:risktraf}

The second stage attaches RiskTraf to the trained backbone obtained in Stage I. Although the Stage-I backbone has access to all three variables, its ERM objective may exploit regime-specific correlations between speed, occupancy, and flow. RiskTraf therefore freezes the backbone prediction $\hat{Y}_{\mathrm{base}}$ and learns only a constrained horizon-wise residual correction, which keeps the plug-in model-agnostic and preserves the spatio-temporal representations learned by the backbone. RiskTraf consists of three components: a zero-start residual head, risk environment construction from historical covariates, and a REx-based objective for robust residual learning. 



\subsubsection{Zero-Start Residual Head}

RiskTraf uses a lightweight residual head to correct the fixed backbone prediction. Since traffic sensors are not exchangeable, the same speed--occupancy pattern may imply different flow corrections at different road segments due to location, lane configuration, detector type, and upstream--downstream context. Motivated by recent studies on spatio-temporal and heterogeneity-aware embeddings~\citep{liu2023stae,dong2024heterogeneity}, RiskTraf introduces a minimal node-identity embedding to provide node-specific context without modifying the frozen backbone.

Specifically, RiskTraf maintains a trainable node-identity embedding table $E=[e_1,\ldots,e_N]^\top \in \mathbb{R}^{N\times d_e}$, where $e_n$ denotes the embedding of sensor node $n$. 
The table is initialized at the beginning of Stage II and optimized together with $\psi$; it is not copied from the backbone and requires no external sensor metadata. In addition, the head uses node-level historical summaries $Q$, including mean speed, mean occupancy, and occupancy-minus-speed over the input window. The residual correction is computed as
\begin{equation}
\Delta \hat{Y} = g_\psi(S_{t-T:t-1}, O_{t-T:t-1}, E, Q),
\end{equation}
where $E$ and $Q$ are broadcast to the batch and horizon dimensions as needed.

The last linear layer of $g_\psi$ is initialized to zero, so RiskTraf initially reproduces the baseline prediction and any nonzero correction must be learned during the debiasing stage.

\subsubsection{Risk Environment Construction}

Risk environments are constructed only from historical speed and occupancy. For each sample in a training batch, we summarize the covariates over the historical window and all nodes:
\begin{align}
\bar{v} &= \frac{1}{TN}\sum_{t,n} v_{t,n}, \quad
\bar{o} = \frac{1}{TN}\sum_{t,n} o_{t,n},
\end{align}
where $T$ is the input length and $N$ is the number of sensor nodes. Inspired by mean alignment~\citep{gao2025stssdl}, we define a normalized traffic-risk score as
\begin{equation}
z(x) = \frac{x-\mu_x}{\sigma_x}, \quad
\rho = z(\bar{o}) - z(\bar{v}),
\end{equation}
where $\mu_x$ and $\sigma_x$ are estimated from the training data. A larger $\rho$ indicates higher occupancy and lower speed, corresponding to more congested or high-risk traffic states. This ordering is grounded in the fundamental diagram of traffic flow, where congestion is characterized jointly by low speed and high occupancy, so $\rho$ ranks samples along the physical free-flow--congestion axis.

Samples are sorted by $\rho$ and partitioned into $K$ ordered environments as follows:
\begin{equation}
e^i = \{(X_j, Y_j) \mid \rho_j \in \text{quantile}_i\}, \quad i = 1, \ldots, K.
\end{equation}
This produces low-to-high traffic-risk environments without requiring incident annotations or manually defined congestion labels.


\subsubsection{REx-Based Residual Learning}

Given the ordered risk environments, RiskTraf trains the residual head to improve flow prediction while keeping residual errors stable across traffic regimes. For each environment $e^i$, we compute
\begin{equation}
\mathcal{L}^{e^i} =
\frac{1}{|e^i|}
\sum_{(X,Y) \in e^i}
\ell(\hat{Y}_{\text{base}} + s_r \cdot g_\psi(S,O, E, Q), Y).
\end{equation}
Let $\bar{\mathcal{L}}=\frac{1}{K}\sum_i \mathcal{L}^{e^i}$. The REx penalty measures the variance of environment losses:
\begin{equation}
\mathcal{R}_{\text{REx}} =
\frac{1}{K}\sum_{i=1}^{K}(\mathcal{L}^{e^i}-\bar{\mathcal{L}})^2.
\end{equation}

The full Stage-II objective is
\begin{equation}
\mathcal{L}_{\text{total}} =
\bar{\mathcal{L}} +
\lambda_\tau \left(
\mathcal{R}_{\text{REx}}
+ \beta \cdot \mathcal{P}_{\text{pair}}
+ \eta \cdot \mathcal{P}_{\text{extra}}
\right),
\end{equation}
where $\beta$ and $\eta$ control the relative weights of the two order-aware penalties, defined on the ordered environment losses as
\begin{equation}
\begin{aligned}
\mathcal{P}_{\text{pair}} &=
\frac{1}{K-1}\sum_{i=1}^{K-1}
\Big[\max\big(0,\, \mathcal{L}^{e^{i+1}}-\mathcal{L}^{e^i}\big)\Big]^2,\\
\mathcal{P}_{\text{extra}} &=
\Big[\max\big(0,\, \mathcal{L}^{e^K}-\operatorname{sg}(\mathcal{L}^{e^1})\big)\Big]^2,
\end{aligned}
\end{equation}
where $\operatorname{sg}(\cdot)$ denotes the stop-gradient operator. $\mathcal{P}_{\text{pair}}$ penalizes any adjacent pair whose higher-risk environment incurs the larger residual loss, so that errors do not grow with the risk level, and $\mathcal{P}_{\text{extra}}$ penalizes the highest-risk loss when it exceeds the detached lowest-risk reference, pushing down only the former. The penalty weight is warmed up as
\begin{equation}
\lambda_\tau =
\lambda \cdot
\min\left(1, \max\left(0, \frac{\tau-\tau_{\text{warmup}}}{\tau_{\text{total}}-\tau_{\text{warmup}}}\right)\right),
\end{equation}
where $\tau$ denotes the current Stage-II epoch. This warmup allows the residual head to first learn useful corrections before enforcing stronger cross-environment consistency.

\subsubsection{Validation Safeguard}
Since RiskTraf is designed as a plug-in correction, it should not degrade a strong backbone. We therefore evaluate the Stage-II checkpoint on the validation set and keep the corrected model only when it improves validation MAE over the frozen baseline. Otherwise, RiskTraf rolls back to the Stage-I prediction. This keeps the plug-in conservative: it improves the backbone when useful residual patterns exist and avoids harmful corrections otherwise.

\subsection{Inference}

At inference, the selected Stage-I backbone first produces the baseline forecast $\hat{Y}_{\text{base}}$. If the Stage-II residual head passes the validation safeguard, RiskTraf computes $\Delta \hat{Y}$ from historical speed and occupancy, node embeddings, and node-level summaries. The final prediction is
\begin{equation}
\hat{Y}_{\text{risk}} = \hat{Y}_{\text{base}} + s_r \cdot \Delta \hat{Y}.
\end{equation}
If the residual head is rolled back, the output remains the Stage-I baseline prediction.

%% file: content/4_experiments.tex
\section{Experiments} \label{sec:experiments}

We conduct experiments on the PEMSB-3V benchmark suite introduced in Section~\ref{sec:benchmark}. The experiments are designed to evaluate whether the proposed benchmark provides physically meaningful three-variable traffic data, and whether RiskTraf can improve traffic flow prediction as a model-agnostic residual plug-in. 
We organize the evaluation around the following research questions:
\begin{itemize}[leftmargin=2em,itemsep=2pt,topsep=2pt]
    \item \textbf{RQ1: Data validity.} Does PEMSB-3V exhibit physically meaningful flow--speed--occupancy relationships?
    \item \textbf{RQ2: Overall effectiveness.} Can RiskTraf consistently improve diverse spatio-temporal forecasting backbones?
    \item \textbf{RQ3: Method comparison.} How does RiskTraf compare with existing debiasing and distribution-shift adaptation methods?
    \item \textbf{RQ4: Component analysis.} How do auxiliary variables, risk penalties, and environment granularity affect RiskTraf?
    \item \textbf{RQ5: Practicality and interpretability.} What overhead does RiskTraf introduce, and what corrections does it learn under ambiguous traffic states?
\end{itemize}

\subsection{Experimental Setup}
\label{subsec:exp_setup}

\paragraph{Datasets and forecasting setting.}
We evaluate RiskTraf on four PEMSB-3V subsets: PEMS03-B, PEMS04-B, PEMS07-B, and PEMS08-B. All experiments follow the same three-variable-to-flow forecasting setting. The input contains 12 historical time steps of flow, speed, and occupancy, corresponding to one hour of observations, and the target contains the next 12 time steps of flow. All metrics are computed on inverse-scaled flow values. We report MAE, RMSE, and MAPE, where lower values indicate better performance.

\paragraph{Backbones.}
To evaluate model-agnostic applicability, we apply RiskTraf to representative spatio-temporal forecasting backbones from different architectural generations, including STGCN~\citep{yu2018spatio}, DCRNN~\citep{li2018dcrnn}, AGCRN~\citep{bai2020adaptive}, Graph WaveNet~\citep{wu2019graphwavenet}, GMAN~\citep{zheng2020gman}, GTS~\citep{shang2021discrete}, STEMGNN~\citep{cao2020spectral}, STNorm~\citep{deng2021st}, STWA~\citep{fang2024efficient}, MegaCRN~\citep{jiang2023megacrn}, HimNet~\citep{dong2024heterogeneity}, and STDN~\citep{Cao_Wang_Jiang_Yu_Dong_2025}. These backbones cover recurrent, graph-based, attention-based, normalization-based, and recent heterogeneity-aware designs.

\paragraph{Paired plug-in protocol.}
For each dataset--backbone pair, we first train a vanilla three-variable backbone using standard empirical risk minimization. RiskTraf then starts from the same validation-selected checkpoint, freezes the backbone, and trains only the lightweight residual head driven by historical speed and occupancy. This paired protocol ensures that any improvement comes from the RiskTraf plug-in rather than from a different backbone initialization or training recipe.

\input{tables/overall.tex}

\paragraph{Implementation details.}
For RiskTraf, each training batch is sorted by the speed--occupancy risk score and partitioned into four ordered traffic-risk environments unless otherwise specified. The residual head is trained with masked MAE and the warmed-up risk extrapolation penalty described in Section~\ref{sub:risktraf}. Baseline backbones and compared methods are implemented using official code when available or reproduced following the original papers, with hyper-parameters selected by validation MAE under the same protocol.
We use Adam for optimization, validation MAE for model selection, and rollback to the Stage-I checkpoint when the plug-in does not improve validation MAE. The batch size is 32 unless otherwise specified. All experiments are conducted on NVIDIA L20 GPUs.

\begin{figure}[tb]
    \centering
    \includegraphics[width=\linewidth]{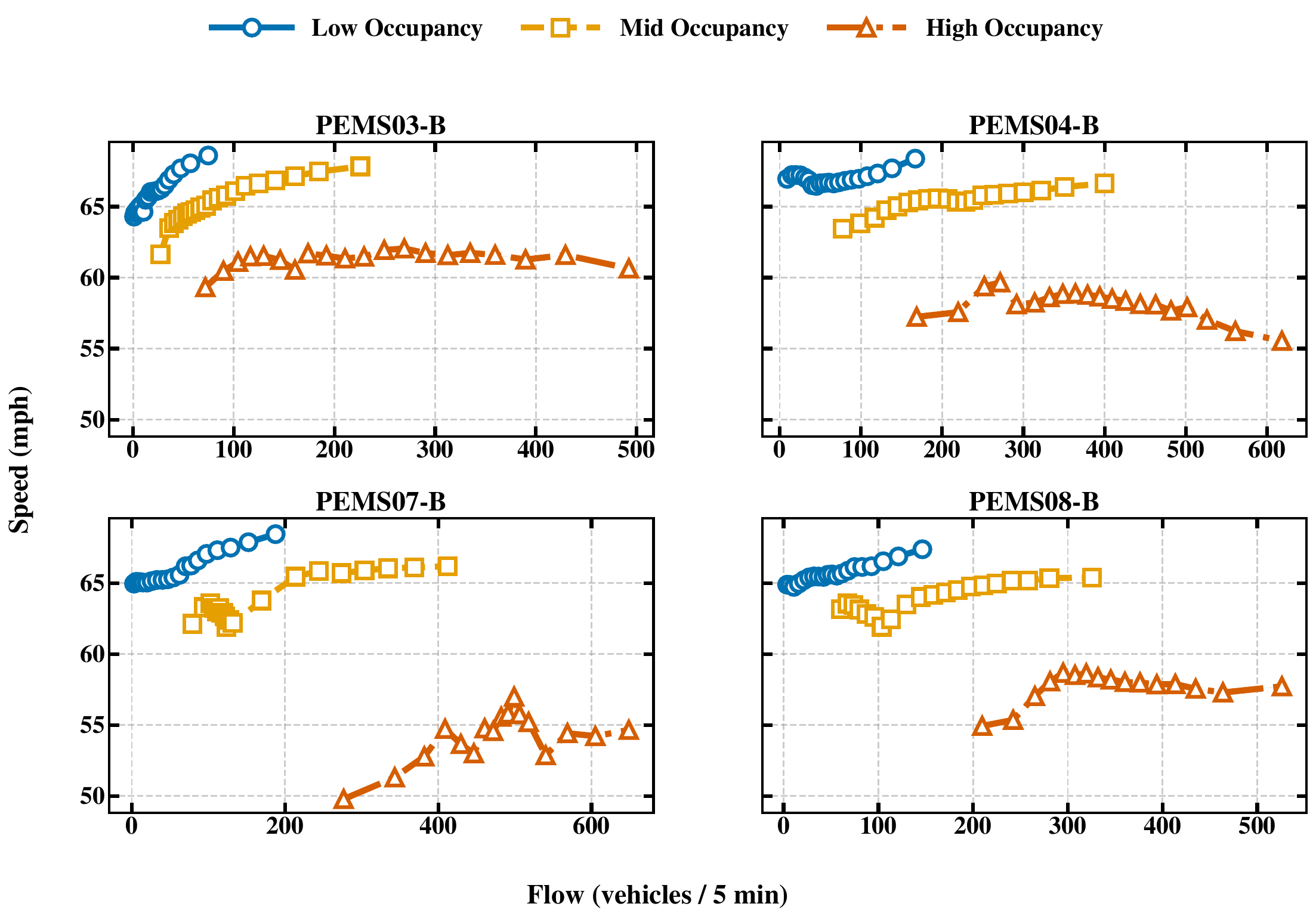}
    \vspace{-1.5em}
    \caption{Flow--speed relationships under different occupancy states on the four PEMSB-3V subsets. Samples are grouped into low-, mid-, and high-occupancy states; each curve shows the average speed at different flow levels.}
    \label{fig:flow_speed_occupancy}
    \Description{Line plots showing that higher occupancy states correspond to lower speeds under similar traffic flow levels on PEMS03-B and PEMS04-B.}
    \vspace{-1.5em}
\end{figure}

\begin{figure*}[th]
    \centering
    \includegraphics[width=.98\textwidth]{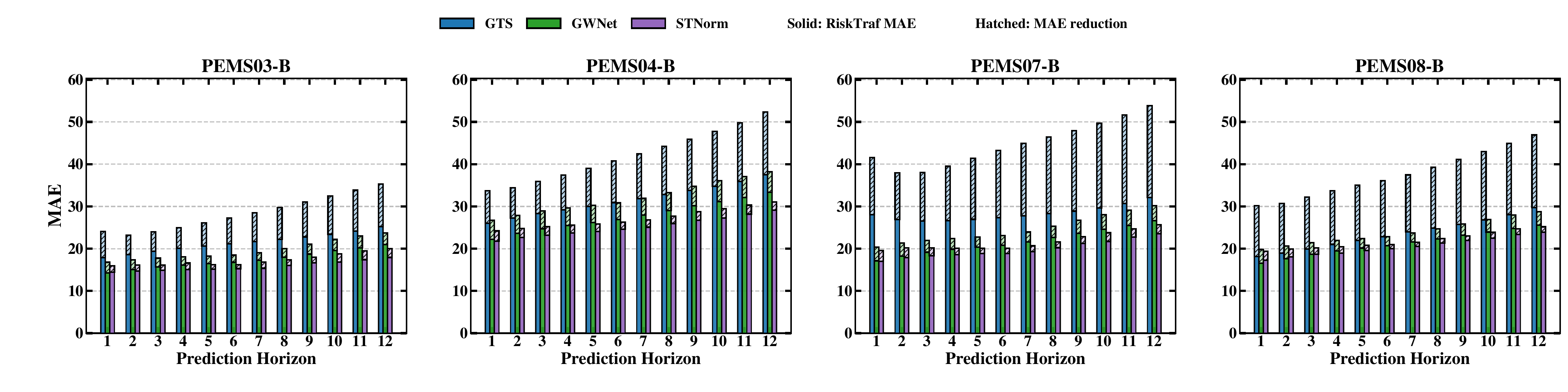}
    \vspace{-1em}
    \caption{Per-horizon MAE decomposition for GTS, GWNet, and STNorm on the four PEMSB-3V subsets. For each horizon, the full bar denotes the baseline MAE, the solid segment denotes the MAE after adding RiskTraf, and the hatched cap denotes the absolute error reduced by RiskTraf.}
    \label{fig:per_step_mae_decomposition}
    \Description{Stacked bar charts for PEMS03-B, PEMS04-B, PEMS07-B, and PEMS08-B showing baseline MAE, RiskTraf MAE, and reduced MAE across prediction horizons 1 to 12 for representative backbones.}
    \vspace{-1em}
\end{figure*}

\subsection{Dataset Characteristics (RQ1)}

We first examine whether PEMSB-3V preserves meaningful relationships among flow, speed, and occupancy. Figure~\ref{fig:flow_speed_occupancy} plots flow--speed curves under low-, mid-, and high-occupancy states. Across the four displayed subsets, higher occupancy generally corresponds to lower speed, which is consistent with the loop-detector mechanism that vehicles occupy detectors for longer durations under congested conditions. For the same flow level, speed can vary substantially across occupancy states, indicating that flow alone is insufficient to identify the underlying traffic regime.

The figure also shows that occupancy is not a trivial proxy for flow. High-occupancy states span a wide range of flow levels and exhibit distinct speed--flow patterns, rather than simply corresponding to the highest flow values. Moreover, the slope and shape of the flow--speed curves differ across occupancy groups and benchmark subsets, suggesting regime-dependent auxiliary correlations. These observations support the use of speed and occupancy as informative traffic-state indicators, while motivating RiskTraf to exploit them through risk-aware residual learning rather than direct unconstrained fusion.




\subsection{Overall Performance (RQ2)}

Table~\ref{overall_performance} reports MAE, RMSE, and MAPE averaged over horizons 3, 6, and 12. For each backbone, the vanilla row denotes the Stage-I model trained with historical flow, speed, and occupancy, while the +RiskTraf row denotes the paired residual plug-in result. 

The results lead to the following observations.

\begin{itemize}[leftmargin=2em,itemsep=2pt,topsep=2pt]
    \item \textbf{RiskTraf consistently improves heterogeneous backbones.} Across the displayed backbones and PEMSB-3V subsets, adding RiskTraf reduces MAE and RMSE in nearly all cases, supporting its model-agnostic plug-in design.

    \item \textbf{Larger gains appear on backbones more sensitive to shifted auxiliary correlations.} DCRNN~\citep{li2018dcrnn} and GTS~\citep{shang2021discrete} obtain particularly large improvements, suggesting that risk-aware residual learning is especially useful when direct three-variable ERM is unstable.

    \item \textbf{Strong recent backbones still benefit from the plug-in.} Competitive models such as STWA~\citep{fang2024efficient}, MegaCRN~\citep{jiang2023megacrn}, HimNet~\citep{dong2024heterogeneity}, and STDN~\citep{Cao_Wang_Jiang_Yu_Dong_2025} also improve with RiskTraf, showing that the plug-in complements rather than replaces advanced backbone architectures.

    \item \textbf{MAPE degradation reflects denominator sensitivity.} A few worse MAPE results do not contradict the MAE/RMSE gains, since MAPE is unstable for zero or near-zero actual values~\citep{HYNDMAN2006679} and behaves like MAE weighted by inverse target magnitude~\citep{DEMYTTENAERE201638}. Thus, better abnormal-regime corrections can still be penalized by small-flow denominators.

\end{itemize}

Figure~\ref{fig:per_step_mae_decomposition} further examines whether the improvements persist at individual prediction horizons. Each bar decomposes the baseline MAE into the remaining error after applying RiskTraf and the absolute error reduced by the plug-in, where the hatched cap directly represents the removed MAE. Across all four subsets, RiskTraf reduces errors on most horizons, showing that the gains are not caused by averaging a few favorable steps. The improvement is especially visible for GTS~\citep{shang2021discrete}, while GWNet~\citep{wu2019graphwavenet} and STNorm~\citep{deng2021st} obtain smaller but still consistent reductions. This suggests that RiskTraf provides larger corrections for backbones with larger horizon-wise errors, while still refining more stable backbones.

The reduction pattern is also horizon-dependent. On PEMS04-B, PEMS07-B, and PEMS08-B, the hatched caps often become more pronounced from short to medium and long horizons, although some subsets also show clear short-horizon corrections when the baseline is unstable. This pattern is consistent with the increasing uncertainty of long-range forecasting: near-term flow is constrained by recent observations, whereas later horizons require recognizing the underlying traffic regime from speed and occupancy. Therefore, the per-horizon decomposition supports that RiskTraf does not simply apply a uniform offset, but learns residual corrections that are more useful when regime-dependent uncertainty accumulates.

\input{tables/debiasing.tex}

\subsection{Comparison with Robust Forecasting Methods (RQ3)}

\input{tables/ablation.tex}

To answer RQ3, we compare RiskTraf with representative robust traffic forecasting methods, including debiasing-based methods CauSTG~\citep{10.1145/3580305.3599421} and ST-SSDL~\citep{gao2025stssdl}, and distribution-shift adaptation methods Dish-TS~\citep{fan2023dishts} and STEVE~\citep{ji2025steve}. Since these methods are standalone forecasting frameworks rather than backbone models, Table~\ref{tab:debiasing_comparison} reports their all-horizon MAE/RMSE separately from the paired plug-in comparison in Table~\ref{overall_performance}. We include STDN~\citep{Cao_Wang_Jiang_Yu_Dong_2025}+RiskTraf as the representative RiskTraf configuration, as STDN is the strongest backbone in the main comparison.

RiskTraf obtains the best MAE and RMSE on all four PEMSB-3V subsets. Compared with the strongest standalone competitor on each subset, it reduces MAE by 8.75\%, 5.12\%, 13.68\%, and 6.09\%, and reduces RMSE by 7.99\%, 3.51\%, 7.87\%, and 1.04\% on PEMS03-B, PEMS04-B, PEMS07-B, and PEMS08-B, respectively. The consistent improvements on both MAE and RMSE indicate that RiskTraf reduces not only average prediction bias but also larger forecasting deviations. Moreover, existing robust methods exhibit dataset-dependent rankings, suggesting that their robustness mechanisms may be sensitive to district-level traffic dynamics. In contrast, RiskTraf remains consistently strong across subsets while preserving the original backbone architecture. This supports the effectiveness of risk-aware residual correction for handling regime-dependent auxiliary correlations without replacing the forecasting model.


\subsection{Component and Sensitivity Analysis (RQ4)}

\paragraph{\textbf{Auxiliary-variable ablation.}}
Table~\ref{tab:ablation} ablates the auxiliary measurements used for traffic-risk environment construction while keeping the backbone, residual head, and training protocol unchanged. The \textit{w/o speed} variant constructs environments using occupancy only, the \textit{w/o speed\&occ} variant removes both speed and occupancy from environment partitioning, and the full variant uses both measurements.

The results show that occupancy alone already provides useful traffic-state information: \textit{w/o speed} consistently outperforms \textit{w/o speed\&occ} on most backbone--dataset pairs. This is expected because occupancy directly reflects detector occupation intensity and is closely related to congestion states. However, the full \textit{speed+occ} variant achieves the best overall performance across all backbone blocks, showing that speed provides complementary information about vehicle movement dynamics. The improvement is particularly clear for DCRNN, while stronger backbones such as GWNet and STWA obtain smaller but still consistent gains. These results confirm that RiskTraf benefits from constructing traffic-risk environments with both auxiliary measurements.

\begin{figure}[t]
    \centering
    \includegraphics[width=0.45\textwidth]{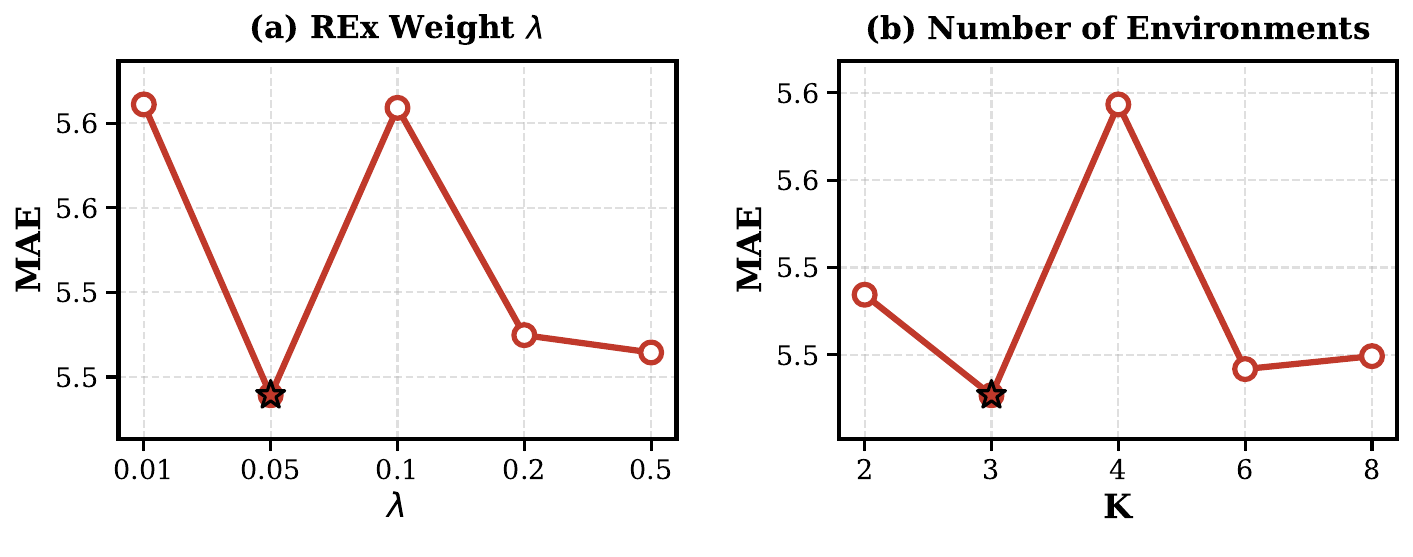}
    \vspace{-1em}
    \caption{Sensitivity analysis of RiskTraf on PEMS03-B with respect to the risk penalty weight $\lambda$ and the number of risk environments $K$. Stars mark the lowest MAE in each sweep.}
    \label{fig:hyper_parameter_study}
    \vspace{-1em}
\end{figure}

\paragraph{\textbf{Sensitivity to $\lambda$ and $K$.}}
Figure~\ref{fig:hyper_parameter_study} studies the sensitivity of RiskTraf on PEMS03-B by varying the risk penalty weight $\lambda$ and the number of environments $K$, while keeping $\beta$ and $\eta$ fixed. The weight $\lambda$ controls the strength of the REx, pairwise, and extrapolation regularizers. A small $\lambda$ weakens cross-environment constraints, whereas an overly large $\lambda$ may sacrifice pointwise accuracy for environment consistency. The best performance is achieved around $\lambda=0.05$.

The environment number $K$ controls the granularity of traffic-risk partitioning. The non-monotonic curve shows that environment construction is not simply improved by using more partitions. Although $K=3$ obtains the lowest MAE in this PEMS03-B sweep, we use $K=4$ in the main experiments as a fixed quartile split for consistent evaluation across all dataset--backbone pairs. This avoids tuning $K$ separately for each setting.

\begin{figure}[t]
    \centering
    \includegraphics[width=\linewidth]{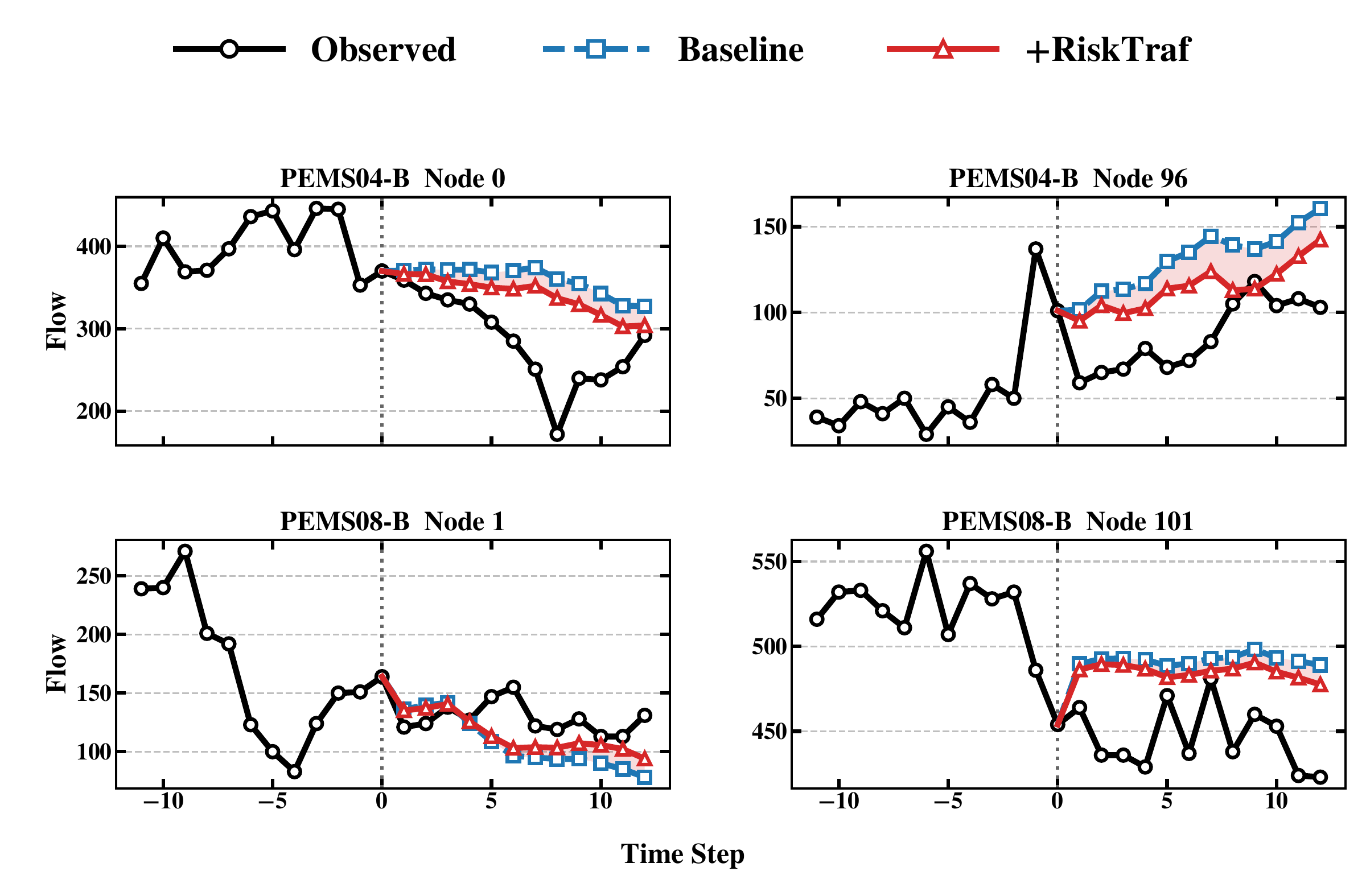}
    \vspace{-1em}
    \caption{Targeted STAEformer forecasting cases on PEMS04-B and PEMS08-B. The black curve denotes observed flow, the red dashed curve denotes the baseline prediction, and the blue curve denotes the prediction after adding RiskTraf. The vertical dashed line marks the prediction start, and the shaded region highlights the residual correction.}
    \label{fig:staeformer_targeted_cases}
    \vspace{-1em}
\end{figure}

\subsection{Efficiency Study (RQ5)}
\input{tables/efficiency.tex}

Table~\ref{tab:efficiency} evaluates RiskTraf from a plug-in efficiency perspective. Since RiskTraf is not designed as a standalone forecasting architecture, the relevant comparison is between each backbone and the same backbone equipped with RiskTraf. Across GTS~\citep{shang2021discrete}, GWNet~\citep{wu2019graphwavenet}, and STNorm~\citep{deng2021st} on PEMS03-B, RiskTraf introduces only about 0.004M additional trainable parameters and marginal GFLOP increases. The inference latency and inference memory remain in the same practical range, indicating that the residual head adds limited deployment overhead.

The small overhead is accompanied by clear accuracy gains. Test MAE decreases from 28.45 to 21.45 for GTS, from 19.65 to 17.43 for GWNet, and from 17.31 to 15.78 for STNorm. These improvements show that the auxiliary risk environments and residual correction provide a favorable accuracy--efficiency trade-off. Training wall-clock time is reported as an auxiliary measurement, since it is affected by Stage-II early stopping and by the fact that only the residual head is optimized while the backbone is frozen.



\begin{figure}[t]
    \centering
    \includegraphics[width=0.7\linewidth]{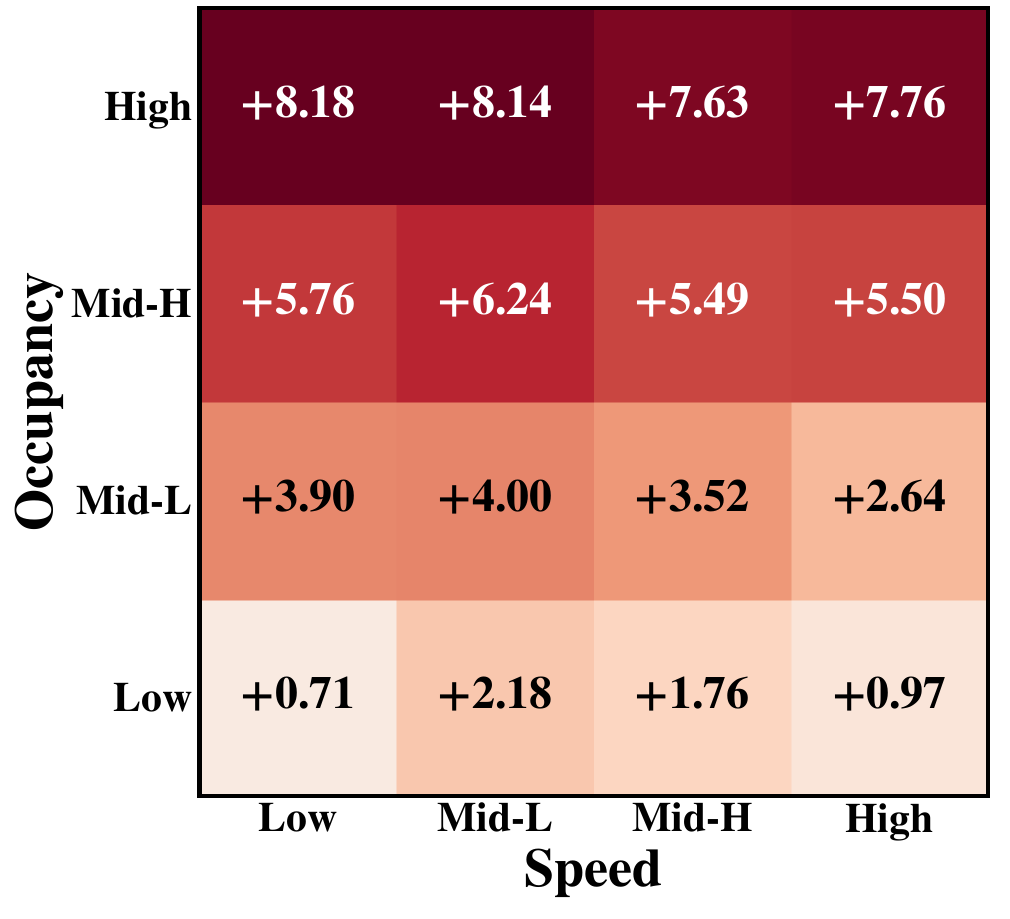}
    \vspace{-1em}
    \caption{RiskTraf correction over STAEformer on PEMS03-B. Samples are grouped by speed and occupancy bins; each cell reports $\mathrm{mean}(\hat{Y}_{\mathrm{RiskTraf}}-\hat{Y}_{\mathrm{Base}})$, where larger positive values indicate stronger upward residual corrections.}
    \label{fig:speed_occ_correction_heatmap}
    \vspace{-1em}
\end{figure}
    
\subsection{Case Study (RQ5)}

\paragraph{Trajectory-level correction.}
Figure~\ref{fig:staeformer_targeted_cases} visualizes representative STAEformer forecasting cases on PEMS04-B and PEMS08-B. The baseline sometimes over-reacts or under-reacts after the prediction start, leading to trajectories that deviate from the observed flow. RiskTraf adjusts the baseline prediction toward the observed trajectory in these cases, especially around turning points or short-term trend changes after the prediction boundary. This shows that the residual head does not simply apply a constant shift, but can produce horizon-wise corrections that reshape the forecast trajectory.

\paragraph{State-dependent correction.}
Figure~\ref{fig:speed_occ_correction_heatmap} further groups samples by speed and occupancy states and reports the average correction 
$\mathrm{mean}(\hat{Y}_{\mathrm{RiskTraf}}-\hat{Y}_{\mathrm{Base}})$. The correction is positive in all bins, indicating that RiskTraf generally raises the STAEformer baseline in the selected PEMS03-B setting. More importantly, the correction magnitude is strongly associated with occupancy: low-occupancy states receive relatively small corrections, whereas high-occupancy states receive much larger corrections. In contrast, the variation along the speed axis is less monotonic. This pattern suggests that occupancy provides a stronger congestion-state signal for the residual head, while speed helps refine the correction within each occupancy regime. 

\begin{figure}[t]
    \centering
    \includegraphics[width=0.3\textwidth]{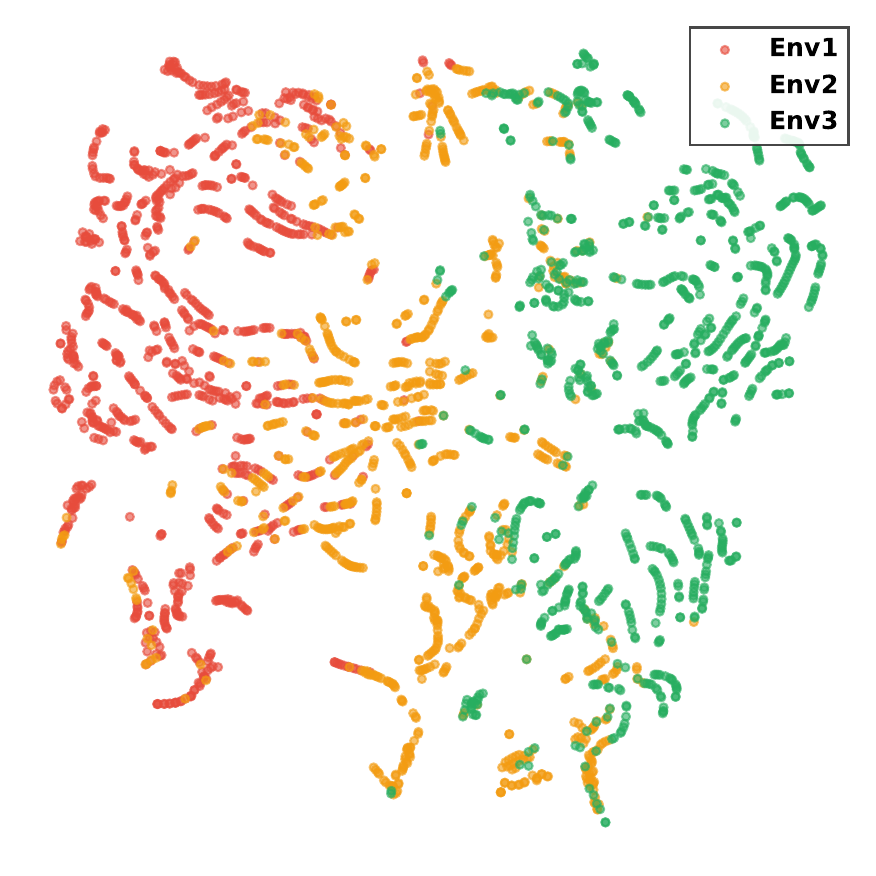}
    \vspace{-1em}
    \caption{t-SNE visualization of PEMS03-B samples colored by RiskTraf risk environments. The three colors denote ordered environments constructed from historical speed and occupancy, showing separated regimes with overlapping transition regions.}
    \label{fig:rex_env_tsne}
    \vspace{-1em}
\end{figure}

\paragraph{Risk-environment structure.}
Figure~\ref{fig:rex_env_tsne} visualizes the risk-environment partition on PEMS03-B using t-SNE. The three environments form a coarse low-to-high risk organization with visible separation, while still retaining overlap in transition regions. The separated regions suggest that the speed--occupancy risk score captures distinct traffic regimes, whereas the overlapping regions correspond to samples whose raw traffic patterns are less clearly separable. This structure is consistent with the motivation of REx-based residual learning: RiskTraf does not require perfectly separated environments, but encourages the residual head to remain stable across related risk regimes while still adapting to high-risk states.

Overall, these case studies provide qualitative evidence for the mechanism behind RiskTraf. The plug-in changes individual trajectories, applies larger corrections under high-occupancy traffic states, and relies on risk environments that reflect meaningful structure in the raw traffic data.

%% file: tables/overall.tex
\begin{table*}[t]
    \caption{Main plug-in results on PEMSB-3V, averaged over horizons 3, 6, and 12. Lower is better; MAPE is in percentage form. Green/red arrows mark improvement/degradation over the backbone row.}
    \label{overall_performance}
    \vspace{-1em}
    \definecolor{riskgreen}{RGB}{0,96,64}
    \newcommand{\basearrow}{\phantom{$\uparrow$}}
    \renewcommand{\arraystretch}{0.76}
    \centering
    \scriptsize
    \setlength{\tabcolsep}{3.6pt}
    \resizebox{\textwidth}{!}{
    \begin{tabular}{l|ccc|ccc|ccc|ccc}
        \toprule
        \multirow{2}{*}{Method} & \multicolumn{3}{c|}{PEMS03-B} & \multicolumn{3}{c|}{PEMS04-B} & \multicolumn{3}{c|}{PEMS07-B} & \multicolumn{3}{c}{PEMS08-B} \\
        \cmidrule(lr){2-4} \cmidrule(lr){5-7} \cmidrule(lr){8-10} \cmidrule(lr){11-13}
        & MAE$\downarrow$ & RMSE$\downarrow$ & MAPE(\%)$\downarrow$ & MAE$\downarrow$ & RMSE$\downarrow$ & MAPE(\%)$\downarrow$ & MAE$\downarrow$ & RMSE$\downarrow$ & MAPE(\%)$\downarrow$ & MAE$\downarrow$ & RMSE$\downarrow$ & MAPE(\%)$\downarrow$ \\
        \midrule
        STGCN & 16.30\basearrow & 27.04\basearrow & 31.20\basearrow & 24.73\basearrow & 37.94\basearrow & 14.70\basearrow & 19.66\basearrow & 32.95\basearrow & 10.81\basearrow & 19.43\basearrow & 32.87\basearrow & 19.06\basearrow \\
        +RiskTraf & 15.86{\color{riskgreen}$\uparrow$} & 26.27{\color{riskgreen}$\uparrow$} & 28.27{\color{riskgreen}$\uparrow$} & 24.23{\color{riskgreen}$\uparrow$} & 37.32{\color{riskgreen}$\uparrow$} & 13.77{\color{riskgreen}$\uparrow$} & 18.66{\color{riskgreen}$\uparrow$} & 32.29{\color{riskgreen}$\uparrow$} & 9.57{\color{riskgreen}$\uparrow$} & 18.75{\color{riskgreen}$\uparrow$} & 32.13{\color{riskgreen}$\uparrow$} & 17.86{\color{riskgreen}$\uparrow$} \\
        \cmidrule(lr){1-13}
        DCRNN & 47.38\basearrow & 71.99\basearrow & 203.44\basearrow & 56.39\basearrow & 80.63\basearrow & 41.98\basearrow & 110.81\basearrow & 141.90\basearrow & 141.00\basearrow & 53.82\basearrow & 77.69\basearrow & 78.09\basearrow \\
        +RiskTraf & 25.72{\color{riskgreen}$\uparrow$} & 39.23{\color{riskgreen}$\uparrow$} & 140.30{\color{riskgreen}$\uparrow$} & 30.93{\color{riskgreen}$\uparrow$} & 45.61{\color{riskgreen}$\uparrow$} & 19.03{\color{riskgreen}$\uparrow$} & 25.59{\color{riskgreen}$\uparrow$} & 41.89{\color{riskgreen}$\uparrow$} & 13.37{\color{riskgreen}$\uparrow$} & 25.80{\color{riskgreen}$\uparrow$} & 39.82{\color{riskgreen}$\uparrow$} & 29.89{\color{riskgreen}$\uparrow$} \\
        \cmidrule(lr){1-13}
        AGCRN & 17.24\basearrow & 29.39\basearrow & 25.17\basearrow & 25.71\basearrow & 39.59\basearrow & 14.46\basearrow & 20.03\basearrow & 33.78\basearrow & 9.10\basearrow & 19.96\basearrow & 33.76\basearrow & 17.81\basearrow \\
        +RiskTraf & 16.61{\color{riskgreen}$\uparrow$} & 27.65{\color{riskgreen}$\uparrow$} & 24.45{\color{riskgreen}$\uparrow$} & 25.27{\color{riskgreen}$\uparrow$} & 38.84{\color{riskgreen}$\uparrow$} & 14.17{\color{riskgreen}$\uparrow$} & 19.05{\color{riskgreen}$\uparrow$} & 32.77{\color{riskgreen}$\uparrow$} & 8.62{\color{riskgreen}$\uparrow$} & 19.31{\color{riskgreen}$\uparrow$} & 32.92{\color{riskgreen}$\uparrow$} & 17.14{\color{riskgreen}$\uparrow$} \\
        \cmidrule(lr){1-13}
        GWNet & 19.65\basearrow & 33.05\basearrow & 25.05\basearrow & 32.13\basearrow & 48.89\basearrow & 17.34\basearrow & 24.61\basearrow & 41.25\basearrow & 10.97\basearrow & 23.92\basearrow & 39.89\basearrow & 19.47\basearrow \\
        +RiskTraf & 17.43{\color{riskgreen}$\uparrow$} & 29.07{\color{riskgreen}$\uparrow$} & 23.92{\color{riskgreen}$\uparrow$} & 27.75{\color{riskgreen}$\uparrow$} & 42.20{\color{riskgreen}$\uparrow$} & 15.35{\color{riskgreen}$\uparrow$} & 21.69{\color{riskgreen}$\uparrow$} & 36.84{\color{riskgreen}$\uparrow$} & 9.91{\color{riskgreen}$\uparrow$} & 21.22{\color{riskgreen}$\uparrow$} & 35.56{\color{riskgreen}$\uparrow$} & 18.22{\color{riskgreen}$\uparrow$} \\
        \cmidrule(lr){1-13}
        GMAN & 14.07\basearrow & 23.68\basearrow & 23.38\basearrow & 23.26\basearrow & 36.00\basearrow & 13.06\basearrow & 16.98\basearrow & 30.65\basearrow & 9.28\basearrow & 17.06\basearrow & 30.55\basearrow & 15.95\basearrow \\
        +RiskTraf & 13.74{\color{riskgreen}$\uparrow$} & 23.46{\color{riskgreen}$\uparrow$} & 23.66{\color{red}$\downarrow$} & 22.12{\color{riskgreen}$\uparrow$} & 34.91{\color{riskgreen}$\uparrow$} & 12.43{\color{riskgreen}$\uparrow$} & 15.94{\color{riskgreen}$\uparrow$} & 29.79{\color{riskgreen}$\uparrow$} & 7.55{\color{riskgreen}$\uparrow$} & 16.43{\color{riskgreen}$\uparrow$} & 29.81{\color{riskgreen}$\uparrow$} & 15.43{\color{riskgreen}$\uparrow$} \\
        \cmidrule(lr){1-13}
        STEMGNN & 16.09\basearrow & 26.58\basearrow & 26.97\basearrow & 26.18\basearrow & 40.31\basearrow & 14.67\basearrow & 45.99\basearrow & 79.86\basearrow & 74.35\basearrow & 20.85\basearrow & 34.59\basearrow & 21.96\basearrow \\
        +RiskTraf & 15.68{\color{riskgreen}$\uparrow$} & 25.80{\color{riskgreen}$\uparrow$} & 26.54{\color{riskgreen}$\uparrow$} & 25.29{\color{riskgreen}$\uparrow$} & 38.99{\color{riskgreen}$\uparrow$} & 14.18{\color{riskgreen}$\uparrow$} & 40.64{\color{riskgreen}$\uparrow$} & 64.59{\color{riskgreen}$\uparrow$} & 59.04{\color{riskgreen}$\uparrow$} & 19.73{\color{riskgreen}$\uparrow$} & 33.02{\color{riskgreen}$\uparrow$} & 19.40{\color{riskgreen}$\uparrow$} \\
        \cmidrule(lr){1-13}
        STNorm & 17.31\basearrow & 28.78\basearrow & 25.07\basearrow & 27.16\basearrow & 41.73\basearrow & 14.89\basearrow & 21.63\basearrow & 37.30\basearrow & 9.78\basearrow & 21.86\basearrow & 36.89\basearrow & 19.93\basearrow \\
        +RiskTraf & 15.78{\color{riskgreen}$\uparrow$} & 26.10{\color{riskgreen}$\uparrow$} & 23.23{\color{riskgreen}$\uparrow$} & 25.20{\color{riskgreen}$\uparrow$} & 38.71{\color{riskgreen}$\uparrow$} & 14.18{\color{riskgreen}$\uparrow$} & 19.87{\color{riskgreen}$\uparrow$} & 34.68{\color{riskgreen}$\uparrow$} & 8.98{\color{riskgreen}$\uparrow$} & 20.51{\color{riskgreen}$\uparrow$} & 34.37{\color{riskgreen}$\uparrow$} & 19.20{\color{riskgreen}$\uparrow$} \\
        \cmidrule(lr){1-13}
        GTS & 28.45\basearrow & 50.08\basearrow & 49.64\basearrow & 41.99\basearrow & 62.02\basearrow & 24.72\basearrow & 44.76\basearrow & 70.24\basearrow & 24.12\basearrow & 37.53\basearrow & 57.01\basearrow & 63.22\basearrow \\
        +RiskTraf & 21.45{\color{riskgreen}$\uparrow$} & 34.87{\color{riskgreen}$\uparrow$} & 46.49{\color{riskgreen}$\uparrow$} & 31.53{\color{riskgreen}$\uparrow$} & 47.18{\color{riskgreen}$\uparrow$} & 17.66{\color{riskgreen}$\uparrow$} & 28.32{\color{riskgreen}$\uparrow$} & 48.27{\color{riskgreen}$\uparrow$} & 16.00{\color{riskgreen}$\uparrow$} & 23.51{\color{riskgreen}$\uparrow$} & 39.50{\color{riskgreen}$\uparrow$} & 36.94{\color{riskgreen}$\uparrow$} \\
        \cmidrule(lr){1-13}
        STWA & 15.38\basearrow & 25.63\basearrow & 21.69\basearrow & 24.74\basearrow & 38.12\basearrow & 14.29\basearrow & 19.29\basearrow & 32.63\basearrow & 10.26\basearrow & 19.85\basearrow & 33.59\basearrow & 17.99\basearrow \\
        +RiskTraf & 14.86{\color{riskgreen}$\uparrow$} & 24.79{\color{riskgreen}$\uparrow$} & 21.06{\color{riskgreen}$\uparrow$} & 24.15{\color{riskgreen}$\uparrow$} & 37.34{\color{riskgreen}$\uparrow$} & 13.66{\color{riskgreen}$\uparrow$} & 18.42{\color{riskgreen}$\uparrow$} & 31.96{\color{riskgreen}$\uparrow$} & 8.36{\color{riskgreen}$\uparrow$} & 18.94{\color{riskgreen}$\uparrow$} & 32.54{\color{riskgreen}$\uparrow$} & 17.47{\color{riskgreen}$\uparrow$} \\
        \cmidrule(lr){1-13}
        MegaCRN & 15.07\basearrow & 25.14\basearrow & 20.33\basearrow & 24.50\basearrow & 37.70\basearrow & 13.16\basearrow & 18.97\basearrow & 33.36\basearrow & 7.97\basearrow & 18.67\basearrow & 32.06\basearrow & 15.24\basearrow \\
        +RiskTraf & 14.89{\color{riskgreen}$\uparrow$} & 24.82{\color{riskgreen}$\uparrow$} & 20.82{\color{red}$\downarrow$} & 23.98{\color{riskgreen}$\uparrow$} & 37.14{\color{riskgreen}$\uparrow$} & 12.99{\color{riskgreen}$\uparrow$} & 18.12{\color{riskgreen}$\uparrow$} & 32.44{\color{riskgreen}$\uparrow$} & 7.61{\color{riskgreen}$\uparrow$} & 18.22{\color{riskgreen}$\uparrow$} & 31.45{\color{riskgreen}$\uparrow$} & 15.43{\color{red}$\downarrow$} \\
        \cmidrule(lr){1-13}
        HimNet & 15.09\basearrow & 25.42\basearrow & 20.92\basearrow & 23.92\basearrow & 37.28\basearrow & 13.28\basearrow & 19.70\basearrow & 33.82\basearrow & 8.78\basearrow & 18.41\basearrow & 31.82\basearrow & 15.63\basearrow \\
        +RiskTraf & 14.72{\color{riskgreen}$\uparrow$} & 24.79{\color{riskgreen}$\uparrow$} & 21.26{\color{red}$\downarrow$} & 23.46{\color{riskgreen}$\uparrow$} & 36.74{\color{riskgreen}$\uparrow$} & 13.10{\color{riskgreen}$\uparrow$} & 18.55{\color{riskgreen}$\uparrow$} & 32.36{\color{riskgreen}$\uparrow$} & 8.44{\color{riskgreen}$\uparrow$} & 17.73{\color{riskgreen}$\uparrow$} & 30.96{\color{riskgreen}$\uparrow$} & 15.57{\color{riskgreen}$\uparrow$} \\
        \cmidrule(lr){1-13}
        STDN & 13.78\basearrow & 22.67\basearrow & 33.80\basearrow & 22.56\basearrow & 35.17\basearrow & 12.54\basearrow & 15.86\basearrow & 29.58\basearrow & 7.36\basearrow & 16.59\basearrow & 30.34\basearrow & 15.61\basearrow \\
        +RiskTraf & 13.24{\color{riskgreen}$\uparrow$} & 22.35{\color{riskgreen}$\uparrow$} & 20.36{\color{riskgreen}$\uparrow$} & 21.85{\color{riskgreen}$\uparrow$} & 34.64{\color{riskgreen}$\uparrow$} & 12.44{\color{riskgreen}$\uparrow$} & 14.95{\color{riskgreen}$\uparrow$} & 29.04{\color{riskgreen}$\uparrow$} & 6.71{\color{riskgreen}$\uparrow$} & 16.04{\color{riskgreen}$\uparrow$} & 29.59{\color{riskgreen}$\uparrow$} & 14.95{\color{riskgreen}$\uparrow$} \\
        \bottomrule
    \end{tabular}
    }
    \vspace{-1em}
\end{table*}

%% file: tables/debiasing.tex
\begin{table}[tb]
    \caption{Comparison with robust traffic forecasting methods on PEMSB-3V. Results are averaged over horizons 3, 6, and 12. The first four methods are standalone robustness methods, while the last column reports RiskTraf plugged into the strongest backbone from Table~\ref{overall_performance}. Best results are in bold.}
    \label{tab:debiasing_comparison}
    \vspace{-1em}
    \renewcommand{\arraystretch}{0.95}
    \centering
    \small
    \setlength{\tabcolsep}{2.0pt}
    \begin{tabular}{ll|ccccc}
        \toprule
        Dataset & Metric & CauSTG & ST-SSDL & Dish-TS & STEVE & \shortstack{RiskTraf (STDN)} \\
        \midrule
        \multirow{2}{*}{PEMS03-B} & MAE$\downarrow$ & 32.82 & 14.94 & 14.51 & 16.95 & \textbf{13.24} \\
        & RMSE$\downarrow$ & 53.95 & 24.91 & 24.29 & 27.24 & \textbf{22.35} \\
        \cmidrule(lr){1-7}
        \multirow{2}{*}{PEMS04-B} & MAE$\downarrow$ & 53.43 & 23.55 & 23.03 & 24.86 & \textbf{21.85} \\
        & RMSE$\downarrow$ & 75.93 & 37.06 & 35.90 & 36.10 & \textbf{34.64} \\
        \cmidrule(lr){1-7}
        \multirow{2}{*}{PEMS07-B} & MAE$\downarrow$ & 62.79 & 17.32 & 17.72 & 20.62 & \textbf{14.95} \\
        & RMSE$\downarrow$ & 86.79 & 31.52 & 33.37 & 31.91 & \textbf{29.04} \\
        \cmidrule(lr){1-7}
        \multirow{2}{*}{PEMS08-B} & MAE$\downarrow$ & 54.41 & 17.08 & 18.47 & 18.77 & \textbf{16.04} \\
        & RMSE$\downarrow$ & 82.58 & 30.71 & 32.25 & 29.90 & \textbf{29.59} \\
        \bottomrule
    \end{tabular}
    \vspace{-2em}
\end{table}

%% file: tables/ablation.tex
\begin{table*}[tb]
    \caption{Ablation study on auxiliary measurements used for traffic-risk environment construction. Results are averaged over horizons 3, 6, and 12; MAPE is reported in percentage form. The best result within each backbone--dataset block is in bold.}
    \label{tab:ablation}
    \vspace{-1em}
    \renewcommand{\arraystretch}{0.86}
    \centering
    \scriptsize
    \setlength{\tabcolsep}{2.5pt}
    \resizebox{\textwidth}{!}{

    \begin{tabular}{ll|ccc|ccc|ccc|ccc}

        \toprule
        \multirow{2}{*}{Backbone} & \multirow{2}{*}{Variant} & \multicolumn{3}{c|}{PEMS03-B} & \multicolumn{3}{c|}{PEMS04-B} & \multicolumn{3}{c|}{PEMS07-B} & \multicolumn{3}{c}{PEMS08-B} \\
        \cmidrule(lr){3-5} \cmidrule(lr){6-8} \cmidrule(lr){9-11} \cmidrule(lr){12-14}
        & & MAE$\downarrow$ & RMSE$\downarrow$ & MAPE(\%)$\downarrow$ & MAE$\downarrow$ & RMSE$\downarrow$ & MAPE(\%)$\downarrow$ & MAE$\downarrow$ & RMSE$\downarrow$ & MAPE(\%)$\downarrow$ & MAE$\downarrow$ & RMSE$\downarrow$ & MAPE(\%)$\downarrow$ \\
        \midrule
        \multirow{3}{*}{DCRNN} & w/o speed & 35.70 & 53.36 & 149.01 & 43.50 & 62.29 & 34.10 & 45.85 & 70.52 & 33.63 & 36.86 & 54.11 & 56.48 \\
        & w/o speed\&occ & 39.45 & 59.49 & 154.12 & 49.01 & 71.24 & 34.47 & 61.43 & 83.21 & 66.09 & 46.54 & 68.41 & 61.30 \\
        & Full (speed+occ) & \textbf{25.72} & \textbf{39.23} & \textbf{140.30} & \textbf{30.93} & \textbf{45.61} & \textbf{19.03} & \textbf{25.59} & \textbf{41.89} & \textbf{13.37} & \textbf{25.80} & \textbf{39.82} & \textbf{29.89} \\
        \cmidrule(lr){1-14}
        \multirow{3}{*}{GWNet} & w/o speed & 19.50 & 32.85 & 24.90 & 31.30 & 47.75 & 17.11 & 23.85 & 40.21 & 10.78 & 23.39 & 39.07 & 19.58 \\
        & w/o speed\&occ & 19.52 & 32.88 & 25.16 & 31.80 & 48.46 & 17.14 & 24.48 & 41.09 & 10.96 & 23.71 & 39.60 & 19.63 \\
        & Full (speed+occ) & \textbf{17.43} & \textbf{29.07} & \textbf{23.92} & \textbf{27.75} & \textbf{42.20} & \textbf{15.35} & \textbf{21.69} & \textbf{36.84} & \textbf{9.91} & \textbf{21.22} & \textbf{35.56} & \textbf{18.22} \\
        \cmidrule(lr){1-14}
        \multirow{3}{*}{STWA} & w/o speed & 15.24 & 25.49 & 21.53 & 24.56 & 37.92 & 13.69 & 18.95 & 32.44 & 8.67 & 19.58 & 33.27 & 18.27 \\
        & w/o speed\&occ & 15.28 & 25.56 & 21.68 & 24.68 & 38.07 & 13.75 & 19.11 & 32.58 & 8.68 & 19.73 & 33.52 & 18.14 \\
        & Full (speed+occ) & \textbf{14.86} & \textbf{24.79} & \textbf{21.06} & \textbf{24.15} & \textbf{37.34} & \textbf{13.66} & \textbf{18.42} & \textbf{31.96} & \textbf{8.36} & \textbf{18.94} & \textbf{32.54} & \textbf{17.47} \\
        \bottomrule
    \end{tabular}
        
    }

   \vspace{-1em}
\end{table*}

%% file: tables/efficiency.tex
\begin{table*}[t]
    \caption{Plug-in efficiency comparison on PEMS03-B with batch size 32. Each pair compares a backbone with the same backbone equipped with RiskTraf. Training time denotes the observed wall-clock time of the recorded run, where RiskTraf trains only the residual head with the backbone frozen.}
    \label{tab:efficiency}
    \vspace{-1em}
    \renewcommand{\arraystretch}{0.9}
    \centering
    \scriptsize
    \setlength{\tabcolsep}{3pt}
    \resizebox{\textwidth}{!}{
    \begin{tabular}{llrrrrrrr}
        \toprule
        \textbf{Backbone} & \textbf{Variant} & \textbf{Params (M)} & \textbf{GFLOPs/Batch} & \textbf{Infer ms/Sample} & \textbf{Infer Mem (MB)} & \textbf{Train Time (s)} & \textbf{Train Mem (MB)} & \textbf{Test MAE} \\
        \midrule
        \multirow{2}{*}{GTS}
        & Base & 8.307 & 21.229 & 0.941 & 231.7 & 18.08 & 962.2 & 28.45 \\
        & +RiskTraf & 8.312 & 21.262 & 1.036 & 239.9 & 6.55 & 274.2 & \textbf{21.45} \\
        \midrule
        \multirow{2}{*}{GWNet}
        & Base & 0.031 & 11.484 & 0.162 & 193.9 & 4.95 & 330.7 & 19.65 \\
        & +RiskTraf & 0.035 & 11.518 & 0.159 & 193.9 & 1.72 & 196.8 & \textbf{17.43} \\
        \midrule
        \multirow{2}{*}{STNorm}
        & Base & 0.042 & 1.685 & 0.069 & 143.1 & 5.27 & 287.8 & 17.31 \\
        & +RiskTraf & 0.046 & 1.719 & 0.078 & 143.2 & 2.42 & 145.9 & \textbf{15.78} \\
        \bottomrule
    \end{tabular}
    }
\end{table*}

%% file: content/2_related_work.tex
\section{Related Work}
\label{sec:related}

\subsection{Traffic Flow Prediction}

Early deep approaches relied on recurrent architectures such as LSTM, which are limited in capturing complex spatial dependencies of road networks; later work turned to graph-based designs: DCRNN~\citep{li2018dcrnn} combines diffusion convolutions with sequence-to-sequence learning, and STGCN~\citep{yu2018spatio} adopts a fully convolutional graph architecture, both relying on predefined adjacency matrices from road topology. Graph WaveNet~\citep{wu2019graphwavenet} instead learns adaptive graphs for latent spatial dependencies, and MegaCRN~\citep{jiang2023megacrn} adds memory and meta-learning for heterogeneous patterns. STAEformer~\citep{liu2023stae} showed that Transformers with spatio-temporal adaptive embeddings perform strongly, STDN~\citep{Cao_Wang_Jiang_Yu_Dong_2025} and HimNet~\citep{dong2024heterogeneity} explored seasonality and heterogeneity, and ST-SSDL~\citep{gao2025stssdl} aligns inputs with historical means to reduce stochastic deviations. As architectural gains become marginal~\citep{shao2024benchmark}, some works turn to richer sources such as event information (EastNet~\citep{Wang_Jiang_Xue_Salim_Song_Shibasaki_2022}) or global-scale multimodal data (Terra~\citep{chen2024terra}), which require additional collection and alignment. In contrast, speed and occupancy are sensor-native measurements collected with flow, yet no existing benchmark systematically studies these raw auxiliary measurements for robust flow prediction; PEMSB-3V fills this gap.

\subsection{Invariant Learning and Causal Inference}

Invariant learning assumes that stable mechanisms remain predictive across environments while spurious correlations vary. IRM~\citep{arjovsky2020invariantriskminimization} seeks representations that admit a shared optimal classifier across environments, IRM Games~\citep{ahuja2020irmgames} recasts this as Nash equilibrium finding, REx~\citep{krueger2021rex} minimizes the variance of risks across environments to improve extrapolation, and Group DRO~\citep{sagawa2020groupdro} minimizes worst-group risk for subgroup robustness. Their effectiveness, however, hinges on how environments are defined: DomainBed~\citep{gulrajani2021domainbed} shows limited gains over empirical risk minimization when partitions do not align with the relevant spurious correlations. This is central in traffic prediction, where environments are not directly given. RiskTraf therefore builds ordered risk environments from historical speed and occupancy---physically interpretable indicators of traffic regimes---and applies REx-style regularization to a residual correction head.

\subsection{Distribution Shift in Time Series}

Non-stationarity between training and test periods is commonly handled by statistical normalization: RevIN~\citep{kim2021reversible} removes instance-level statistics before prediction and restores them afterward, Dish-TS~\citep{fan2023dishts} models shifts through learnable transformations, and Non-stationary Transformers~\citep{liu2022nonstationary} introduce de-stationary attention against over-stationarization, while the multi-order wavelet derivative transform~\citep{wavets} captures evolving non-stationary dynamics in the wavelet domain. These methods target shifts in the statistical or spectral properties of the series itself. The challenge studied here is different: the relationships among flow, speed, and occupancy themselves vary across traffic regimes, so treating speed and occupancy as ordinary covariates may introduce shortcut learning. RiskTraf instead uses them to define traffic-risk environments and learns a constrained residual correction that remains stable across these environments.

%% file: content/7_conclusion.tex
\section{Conclusion and Future Works}
\label{sec:conclusion}

In this paper, we study how to effectively use speed and occupancy for traffic flow prediction. While these sensor-native variables provide useful traffic-state information, direct three-variable training can exploit regime-dependent correlations and hurt generalization. To address this, we propose RiskTraf, a paired plug-in that freezes a trained three-variable backbone and learns a lightweight residual correction from historical speed and occupancy under a REx objective. By constructing traffic-risk environments, RiskTraf performs state-dependent flow correction on the frozen backbone. We also introduce PEMSB-3V, a public multi-variate traffic benchmark with raw flow, speed, and occupancy measurements. Experiments show that RiskTraf consistently improves diverse spatio-temporal backbones and outperforms robust forecasting methods. Future work may explore batch-independent REx objectives and extend RiskTraf to other spatio-temporal prediction tasks with auxiliary measurements.

